\documentclass{svmult}

\usepackage{makeidx}         
\usepackage{graphicx}        
\usepackage{multicol}        
\usepackage[bottom]{footmisc}

\usepackage{amsmath}
\usepackage{amsfonts}
\usepackage{amssymb}
\usepackage{url}
\usepackage[caption=false]{subfig}
\usepackage{tabulary}
\usepackage{booktabs}

\DeclareMathOperator{\sign}{sgn}

\makeindex             

\begin{document}

\title*{Evaluation of Hashing Methods Performance on Binary Feature Descriptors}
\titlerunning{Evaluation of Hashing Methods}
\author{Jacek Komorowski\inst{1}\and
Tomasz Trzcinski\inst{2}}
\institute{Warsaw University of Technology, Warsaw, Poland
\texttt{jacek.komorowski@gmail.com}
\and Warsaw University of Technology, Warsaw, Poland \texttt{t.trzcinski@ii.pw.edu.pl}}
%
%
\maketitle

\begin{abstract}
In this paper we evaluate performance of data-dependent hashing methods on binary data.
The goal is to find a hashing method that can effectively produce lower dimensional binary representation of 512-bit FREAK descriptors. 
A representative sample of recent unsupervised, semi-supervised and supervised hashing methods was experimentally evaluated on large datasets of labelled binary FREAK feature descriptors.
\end{abstract}


\section{Introduction}

This paper presents results of an experimental evaluation of recent data-dependent hashing methods applied to binary feature descriptors. The work was motivated by challenges in development of a real-time structure-from-motion solutions for mobile platforms with limited hardware resources.
One of the key elements in a typical structure-from-motion processing pipeline is feature matching step, where correspondences between features detected on a new image and features found on previously processed images are being sought.
Such correspondences are used to compute camera orientation and build a 3D model of an observed scene.

Binary feature descriptors, such as FREAK \cite{6247715}, are good choice for mobile solutions. 
They can be efficiently computed and are very compact (512 bits for FREAK versus 512 bytes for real-valued SIFT descriptor).
Hamming distance between two binary feature descriptors can be quickly computed using few machine code instructions.
But even if comparison of two binary descriptors is very fast, finding correspondence between thousands of features detected in a new image and millions of features on previously processed images requires significant processing power.
For real-valued descriptors efficient approximate nearest neighbours search methods can be applied, such as FLANN \cite{muja2009fast}.
Unfortunately methods based on clustering do not work well with binary data \cite{Trzcinski20122173}.

In this paper we investigate if hashing methods can be used to reduce dimensionality of 512-bit binary FREAK descriptors to improve feature matching performance and lower storage requirements.
Additionally we want to verify if additional training information, if the form of landmark id linked with each descriptor, can be used to improve accuracy of searching for matching descriptors. 

Authors have not encountered any results of evaluation of hashing methods on binary data. 
In the context of image-based information retrieval, hashing algorithms were evaluated on real-valued descriptors such as GIST or SIFT.
Lack of such results motivated the research described in this paper.

The paper is structured as follows. 
Section \ref{jk:sec:hashing} briefly describes hashing methods.
Section \ref{jk:sec:results} presents results of the experimental evaluation of representative hashing methods on large datasets of binary FREAK feature descriptors.
The last section concludes the article and presents ideas for future research.

\section{Overview of hashing methods}
\label{jk:sec:hashing}

Hashing for similarity search is a very active area of development. Many new hashing methods were proposed in the last few years. A number of surveys  documenting current state-of-the-art was published recently
\cite{DBLP:journals/corr/WangZSSS16}
\cite{WangLKC15}
. 

Two major categories of hashing methods can be distinguished: \emph{data-independent} methods and \emph{data-depended} methods.
\emph{Data-independent} methods, such as Locality Sensitive Hashing (LSH) 
\cite{Indyk:1998:ANN:276698.276876}, do not take into account characteristics of the input data.
As such, they have inferior performance in real-live applications, where input data usually has some intrinsic characteristic which can be exploited.
We focus our attention on data-dependent on \emph{data-depended} methods, also known as \emph{learning to hash}. 
These methods exploit properties of the input data to produce more discriminative and compact binary codes.
Data-dependent approach can be further categorized by the level of an external supervision. Unsupervised methods use techniques as spectral analysis or kernelized random projections to compute affinity-preserving
binary codes. They exploit the structure among a sample of unlabelled data to learn appropriate
embeddings. Semi-supervised or supervised methods exploit additional information from annotated
training data. Additional information is usually given as the similarity matrix or list of pairs of similar
and dissimilar items. Semi-supervised methods assume that explicit similarity information is provided
for only a fraction of an input dataset. Affinity between other elements is inferred from the distance in
the input space.

\emph{Learning to hash} is defined \cite{WangSSJ14} as learning a \textbf{compound hash function}, 
$\mathbf{y} = H(\mathbf{x})$,
mapping an input item $\mathbf{x}$ to a compact binary code $\mathbf{y}$, 
such that  nearest neighbour search in the coding space is efficient and the result is a good approximation of the true nearest search result in the input space.
$K$-bit binary code $\mathbf{y} \in \mathbb{B}^{K}$ for a sample point $\mathbf{x} \in \mathbb{R}^D$ is computed as
$\mathbf{y} = \left[ y_1, \ y_2, \cdots y_K,  \right] = \left[ h_1(\mathbf{x}),  h_2(\mathbf{x}), \cdots,  h_K(\mathbf{x})  \right]$.
Each $h_k$ is a \textbf{binary hash function}, mapping elements from $\mathbb{R}^{D}$ to 
$\mathbb{B} = \left\lbrace 0,1\right\rbrace $
A compound hash function $H = \left[ h_1,h_2,\cdots, h_K \right]$ is an ordered set of binary hash functions computing $K$-bit binary code.

Two most popular choices of a \textbf{hash function} are linear projection and kernel-based.
Linear projection hash functions are in the form:
$y = h(\mathbf{x}) = \sign ( \mathbf{w}^{T} \mathbf{x} + b )$,
where $\mathbf{w} \in \mathbb{R}^D$ is the projection vector and $b$ is the bias.
Kernel-based hash functions are in the form:
$y = h(\mathbf{x}) = \sign
\left( 
\sum_{t=1}^{T}
w_t
K
\left( 
\mathbf{s}_t,
\mathbf{x}
\right) 
+b
\right) 
$,
where 
$K$ is a kernel function,
$\mathbf{s}$ is a set of representative samples that are randomly chosen from the dataset or cluster centres of the dataset and $w_t$ are weights.
Other choices of hash function include spherical functions, Laplacian eigenfunctions, neural networks, decision trees-based and non-parametric functions. 

\section{Evaluation results}
\label{jk:sec:results}

\paragraph{Experiment setup}

The experimental evaluation was conducted on datasets consisting of hundred thousands or more labelled 512-bit FREAK feature descriptors. 
Datasets were created by structure-from-motion application developed in Google Tango project
\footnote{See: \url{https://get.google.com/tango/}}.
Each descriptor is labelled with a corresponding landmark id. 
Descriptors with the same landmark id are projections of the same scene point (landmark) on different images.



Table \ref{jk:table:methods} lists hashing methods evaluated in this paper.
Each method was first trained on the training dataset.
Learned hash functions were applied to the test dataset to generate hash codes of a different length: 32, 64, 128 and 256-bits.
Search precision (\emph{Precision@1}) using the resulting hash codes was evaluated and reported.
\emph{Precision@1} for a dataset is calculated as a mean \emph{precision@1} when searching for nearest neighbours using a linear scan for 20 thousand elements randomly sampled from the dataset. 
\emph{Precision@1} for a sampled element is 1, if its nearest neighbour (based on Hamming distance) in the entire dataset is labelled with the same landmark id. 
Otherwise, precision is 0.

\renewcommand{\arraystretch}{1.1}

\begin{table}
\begin{tabulary}{1.0\textwidth}{@{}LLL@{}}
		\toprule 
		\centering Method &  \centering Class
		 & \centering Hash function
		\tabularnewline
		\midrule
		Spectral Hashing (SH) \cite{NIPS2008_3383} & U &  Laplacian eigenfunction 
		\tabularnewline
	
		Binary Reconstructive Embeddings (BRE) \cite{NIPS2009_3667} & U & linear projection	
	  \tabularnewline

		Unsupervised Sequential Projection Learning Hashing (USPLH) \cite{icml2010_WangKC10} & U & linear projection 	        \tabularnewline

		Iterative Quantization (ITQ) \cite{6296665} & U
		& 
		 linear  projection
		\tabularnewline

		Isotropic Hashing (IsoH) \cite{NIPS2012_4846}  & U
		&  linear  projection 
		\tabularnewline
	
		Density Sensitive Hashing (DSH) \cite{DBLP:journals/corr/abs-1205-2930}  & U
		&  linear projection
		\tabularnewline

		Spherical Hashing (SpH) \cite{6248024} & U & spherical function
		\tabularnewline

		Compressed Hashing \cite{lin2013compressed} & U & kernel-based
		\tabularnewline

		Harmonious Hashing (HamH) \cite{Xu:2013:HH:2540128.2540389}  & U &  kernel-based
		\tabularnewline

		Kernelized Locality Sensitive Hashing (KLSH) \cite{kulis2009kernelized} & U &  kernel-based
		\tabularnewline

		Sequential Projection Learning (SPLH) \cite{WangSSH2012} & SS &
		linear projection
		\tabularnewline

		Bootstrap SequentialProjection Learning -- linear version (BTSPLH) \cite{journals/tkde/WuZCCB13} & SS &
		linear projection
		\tabularnewline

		Fast supervised hashing (FastHash) \cite{FastHash2015Lin} & S &
		boosted decision trees
		\tabularnewline


\bottomrule
\end{tabulary}
\caption{List of evaluated hashing methods. Class: U = unsupervised, SS = semi-supervised, S = supervised.}
\label{jk:table:methods}
\end{table}

\paragraph{Unsupervised hashing methods}

This section contains results of an experimental evaluation of unsupervised hashing methods. 
The evaluation was conducted using publicly available MATLAB implementation \cite{2016arXiv161207545C}.
For brief description of methods evaluated in this section refer to \cite{2016arXiv161207545C}.


Hashing methods were trained using 
a dataset consisting of $1\,267\,346$ FREAK descriptors.
The evaluation was done on a separated
test dataset consisting of $342\,602$ descriptors associated with $40\,704$ landmarks.

\begin{figure*}[t!]
    \centering
	\subfloat[Non-kernel based]{\includegraphics[width= 2.55in]{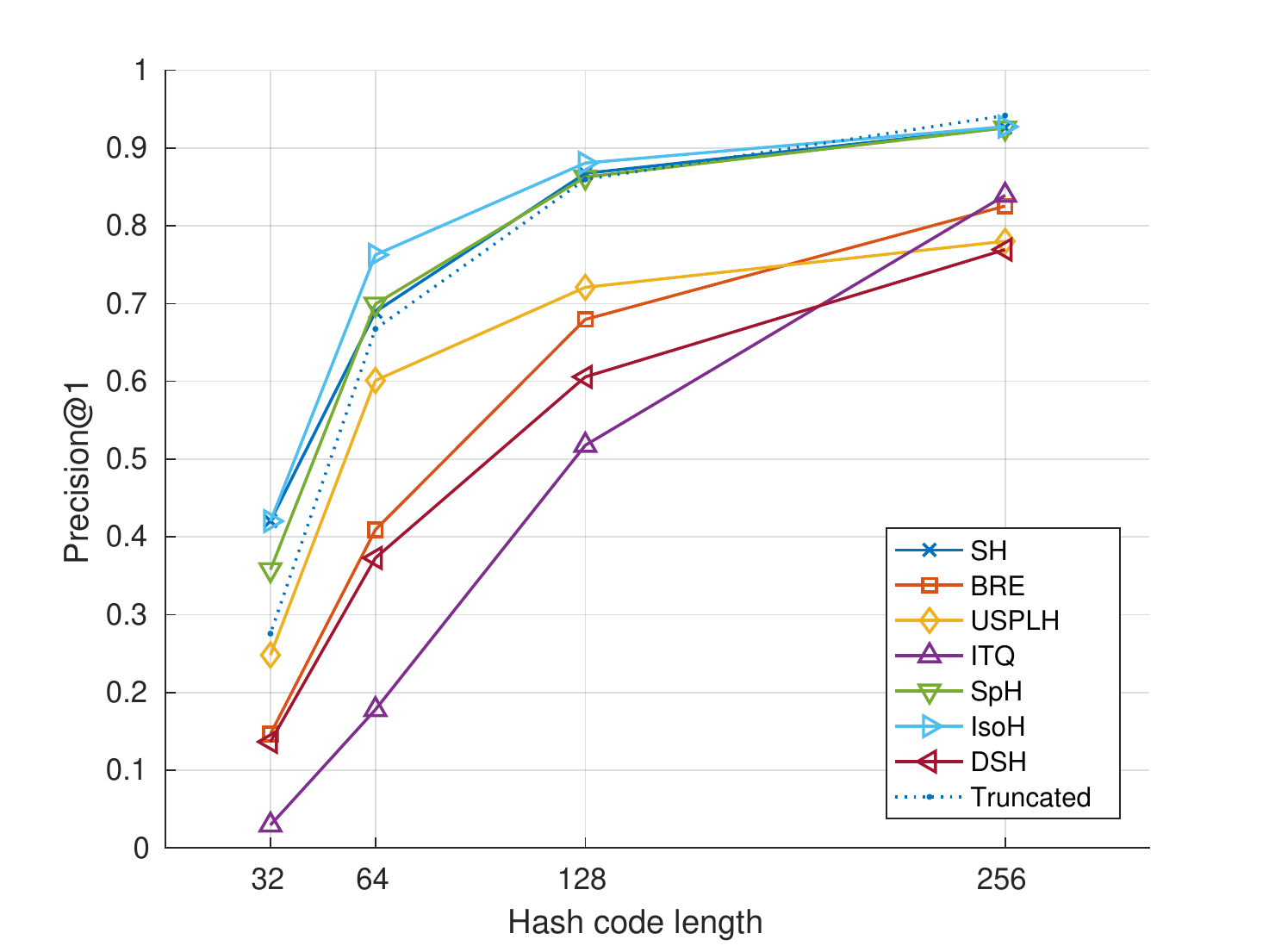}}
	\subfloat[Kernel based methods]{\includegraphics[width= 2.55in]{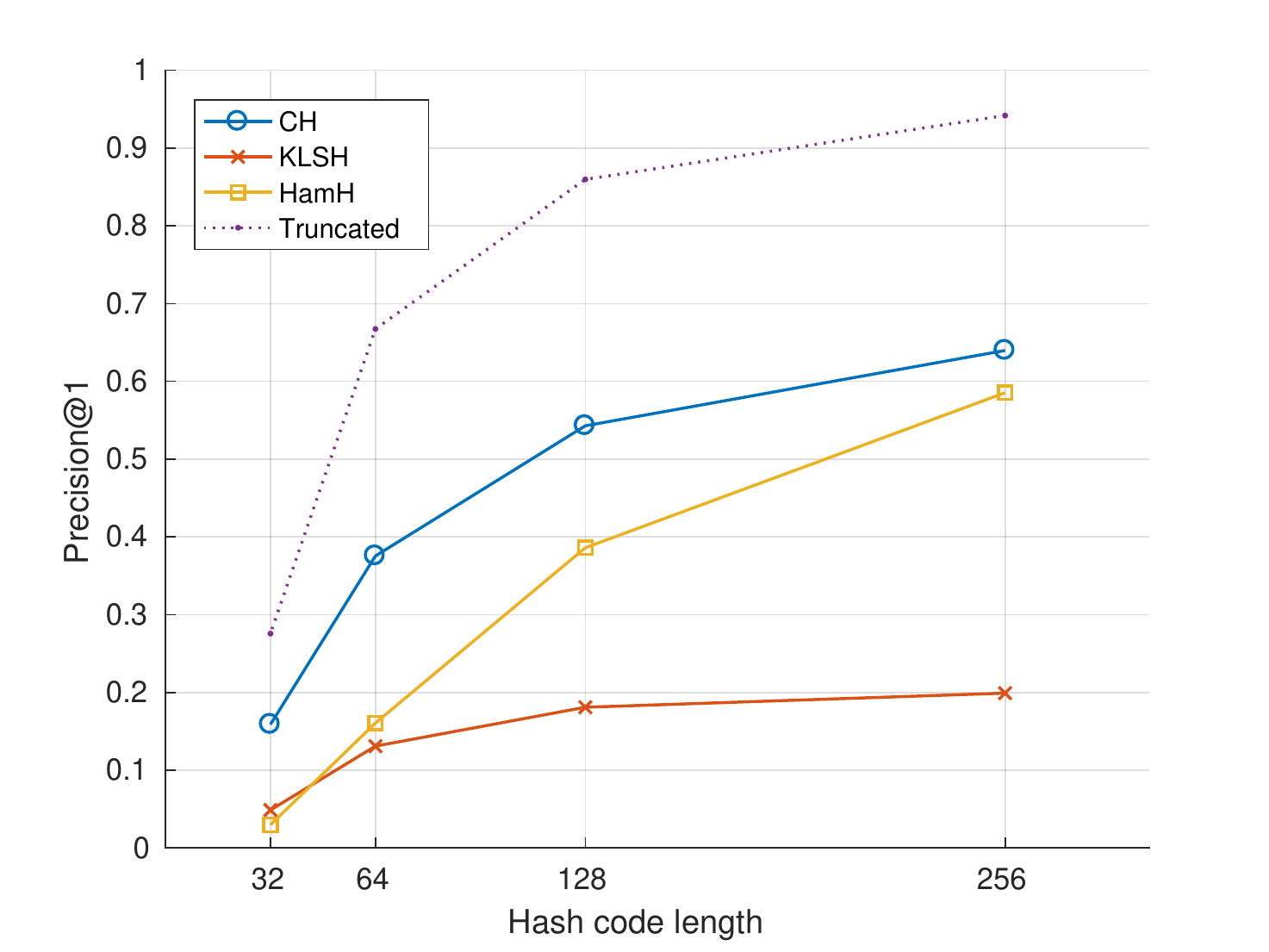}}
	\caption{Search precision using hash codes generated from the test dataset by applying unsupervised hashing methods. Search precision in the test dataset truncated to first $k$ bits is plot with a dotted line for comparison.}
  	\label{jk:fig:anns1}
\end{figure*}

Fig. \ref{jk:fig:anns1} presents search precision in datasets created from the test dataset by applying seven unsupervised, data-dependent hashing methods.
To baseline the results, precision of a linear search on the test dataset truncated to first $k$ bits was evaluated.
It must be noted, that discarding last 256 bits of original FREAK descriptor has little effect on the nearest neighbour search precision. \emph{Precision@1} decreases by $2.1$\% from $0.962$ to $0.942$.
The reason is likely the construction of FREAK descriptor itself, where more discriminative binary tests are used to generate first bits of the descriptor.
Interestingly, for 256-bit hash codes, all evaluated hashing methods perform worse compared to naive bit truncation.
For shorter codes (128-bits and below) only three hashing methods: Isotropic Hashing \cite{NIPS2012_4846} , Spherical Hashing 
\cite{6248024} and Spectral Hashing \cite{NIPS2008_3383} yield better accuracy. 
For 128-bit codes, the advantage of best methods over a naive bit truncation is minimal (1-2\%). 
For the shorter codes best hashing methods perform noticeably better.
Best performing method is Isotropic Hashing (IsoH) \cite{NIPS2012_4846}.

Performance of kernel-based hashing methods on the binary data is very poor. Results are much worse than naive approach of truncating an original dataset to the first $k$ bits.
This is expected as kernel-based hash functions are based on the set of anchor points, that are representative samples or cluster centres for the training dataset.
Clustering methods perform poorly on data from binary spaces. One of the reasons are decision boundaries in binary spaces \cite{Trzcinski20122173}, as large proportion of points in the Hamming space is equidistant from two randomly chosen anchor points.

\paragraph{Semi-supervised and supervised hashing methods}

Semi-supervised and supervised data-dependent hashing methods evaluated in this section were trained using the dataset
consisting of $340\,063$ descriptors associated with $37\,148$ landmarks.
The evaluation was done on 
the separated test dataset consisting of  $342\,602$ descriptors associated with $40\,704$ landmarks.

\begin{figure*}[t!]
    \centering
	\subfloat[$\eta$= 1]{\includegraphics[width= 2.55in]{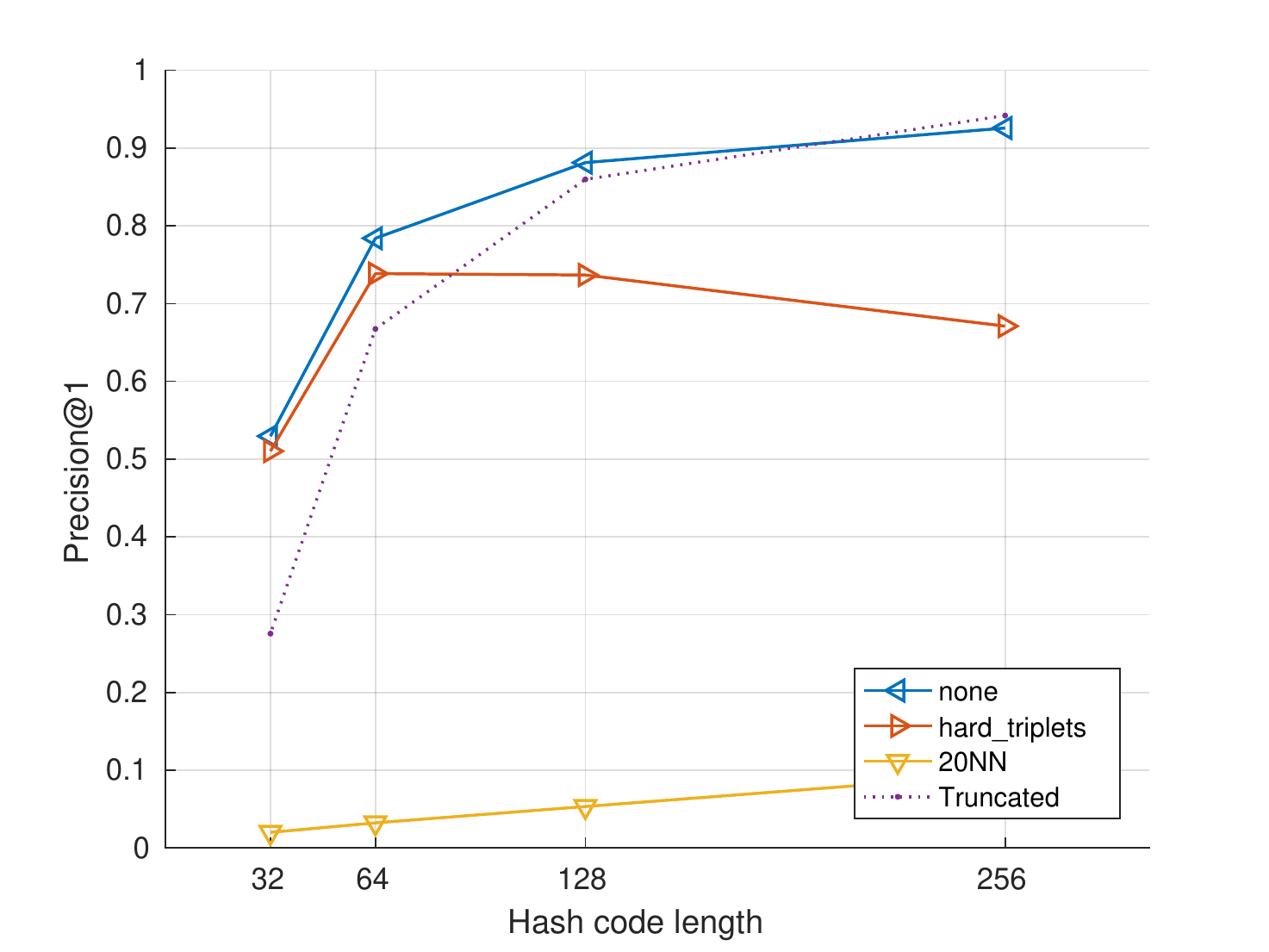}}
	\subfloat[$\eta$ = 100]{\includegraphics[width= 2.55in]{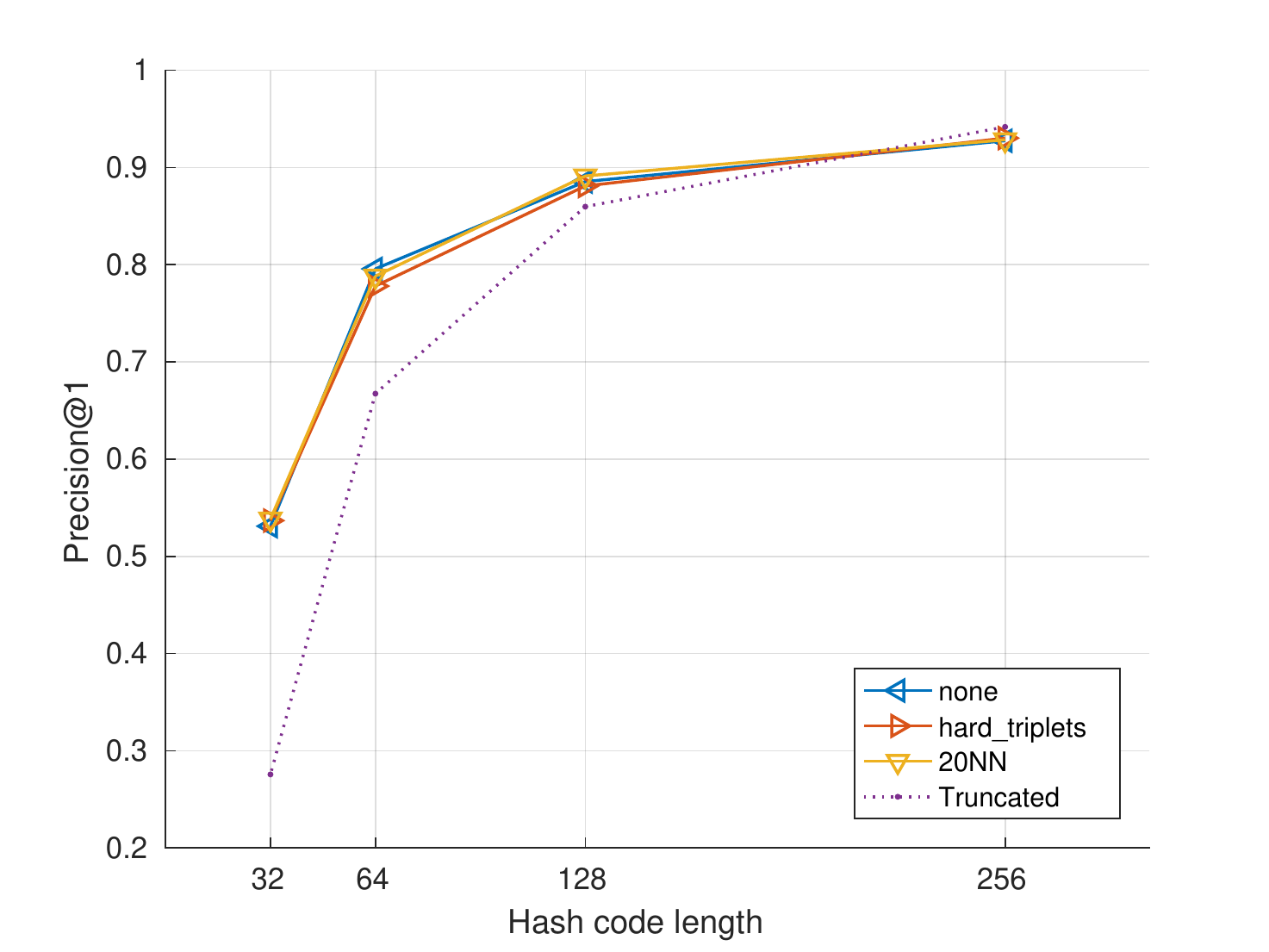}}
	\caption{Search precision using hash codes generated from the test dataset by semi-supervised SPLH method. Different subplots correspond to different weight $\eta$ of an unsupervised term in the objective function.
	On each plot results obtained using three different similarity encodings schemes are shown.
	Search precision in the test dataset truncated to first $k$ bits is plot with a dotted line for comparison.}
  	\label{jk:fig:splh}
\end{figure*}

Figure \ref{jk:fig:splh} presents search precision in datasets created from 
the training dataset by applying semi-supervised Sequential Projection Learning (SPLH) \cite{WangSSH2012} method.
SPLH objective function of consists of two components: supervised empirical fitness and unsupervised information theoretic regularization. A supervised term tries to minimize empirical error on the labelled data. That is, for each bit minimize a number of instances where elements with the same label are
mapped to different values and elements with different labels are mapped to the same value. An unsupervised term provides regularization by maximizing desirable properties like variance and independence of individual
bits.
Different lines on the plot correspond to different similarity encoding schemes.
In \emph{hard triplets} encoding, for each element from the training dataset, the closest element linked with the same landmark id is encoded as similar and the closest element with a different landmark id is encoded as dissimilar.
In \emph{20NN} encoding, for each element from the training dataset, the similarity with its 20 nearest neighbours is encoded.
For comparison, SPLH method was evaluated without any similarity information(\emph{none}), using only unsupervised term in the optimization function.
Surprisingly, using supervised information does not provide any improvement in the search accuracy.
In contrary, when supervised term in the objective function has higher weight (lower $\eta$, left subplot),
the results are noticeably worse, especially for longer codes. The best results are achieved when supervised information is not used at all (\emph{none}).
When unsupervised term in the objective function has higher weight (higher $\eta$, right subplot),
the encoded similarity information has little effect and the performance is the same as without using any supervised information.

\begin{figure*}[t!]
    \centering
	\subfloat[$\lambda$= 1]{\includegraphics[width= 2.55in]{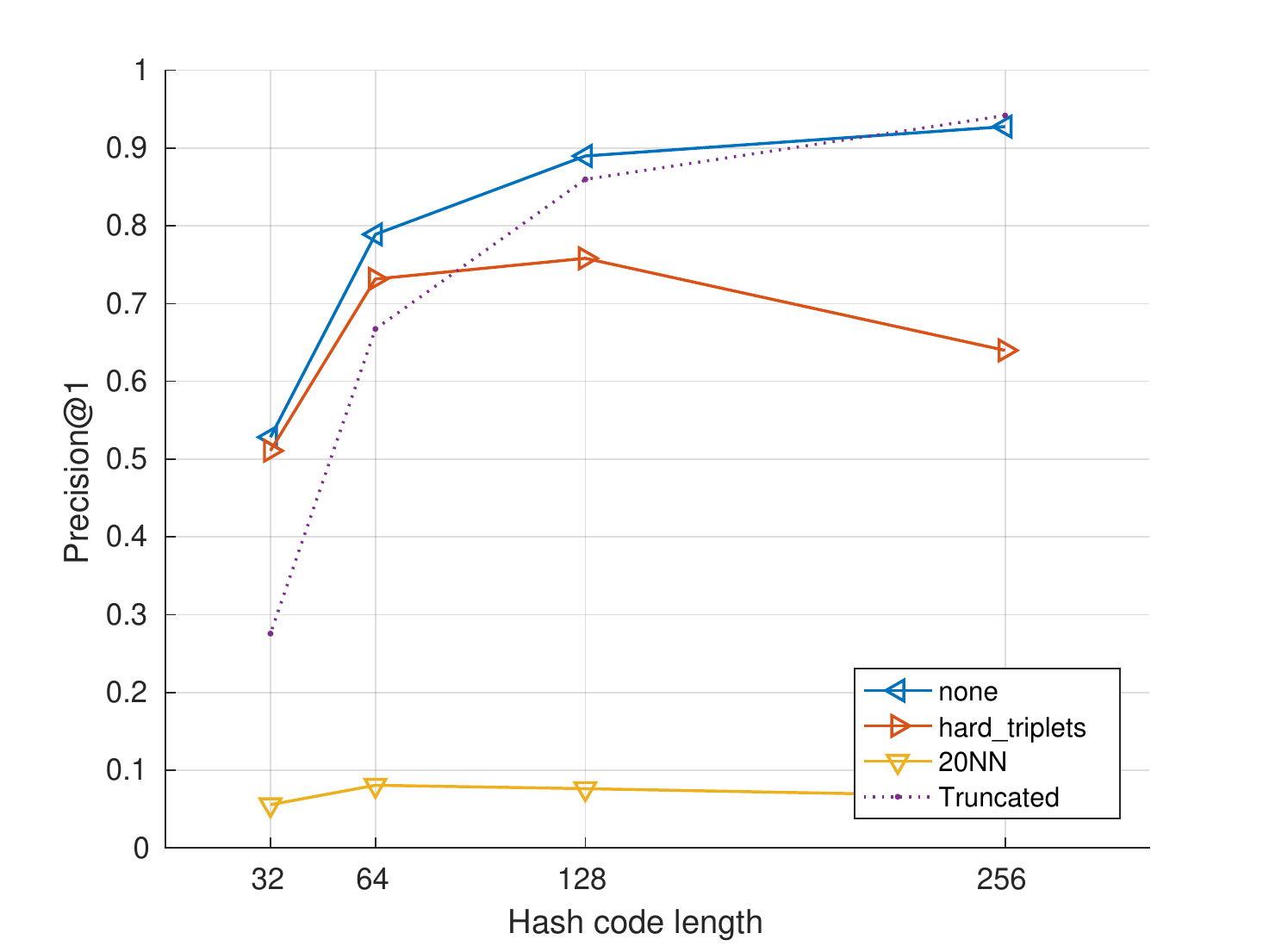}}
	\subfloat[$\lambda$ = 10]{\includegraphics[width= 2.55in]{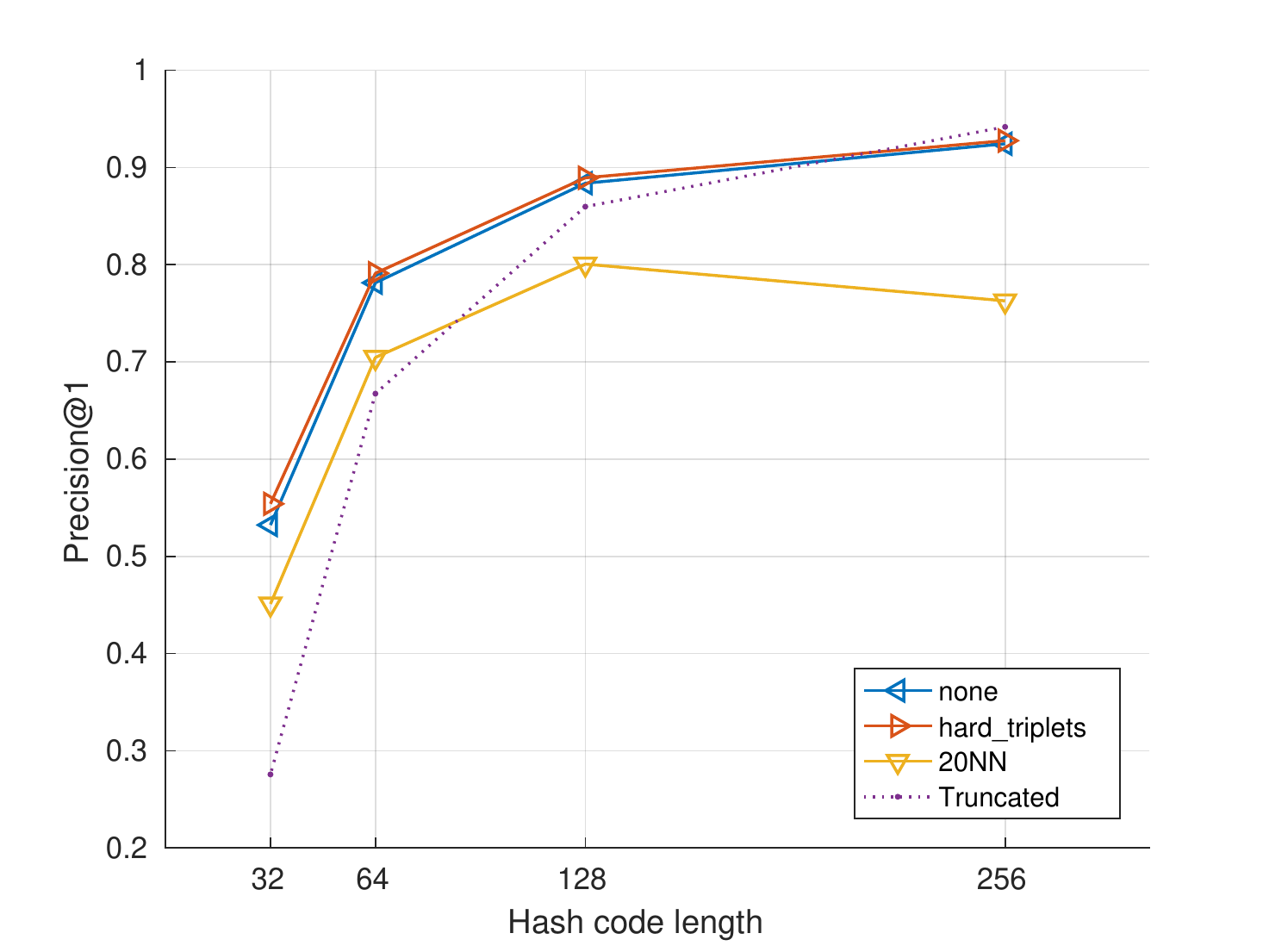}}
	\caption{Search precision using hash codes generated from the test dataset by semi-supervised BTSPLH method. Different subplots correspond to different weight $\lambda$ of an unsupervised term in the objective function.
	On each plot results obtained using three different similarity encodings schemes are shown.
	Search precision in the test dataset truncated to first $k$ bits is plot with a dotted line for comparison.}
  	\label{jk:fig:btsplh}
\end{figure*}

Semi-supervised nonlinear hashing (BTSPLH) \cite{journals/tkde/WuZCCB13} is an enhancement of SPLH method.
Instead of the boosting-like process in SPLH, authors propose a bootstrap-style sequential learning scheme to derive the hash function by correcting the errors incurred by all previously learned bits. 
The results of evaluation of BTSPLH method are depicted on Figure \ref{jk:fig:btsplh}.
The results are very similar to previous SPLH method.
When supervised term in the objective function has higher weight (lower $\lambda$, left subplot),
the results using supervised information (\emph{hard\_triplets} and \emph{20NN} encoding) are poor. The best results are achieved when supervised information is not used at all.

The last method evaluated in this paper is supervised FastHash \cite{lin2015supervised} algorithm.
It uses a two-step learning strategy: binary code inference is followed by binary classification step using an ensemble of decision trees. 

\begin{figure*}[t!]
    \centering
	\subfloat[256-bit codes][]{\includegraphics[width= 2.55in]{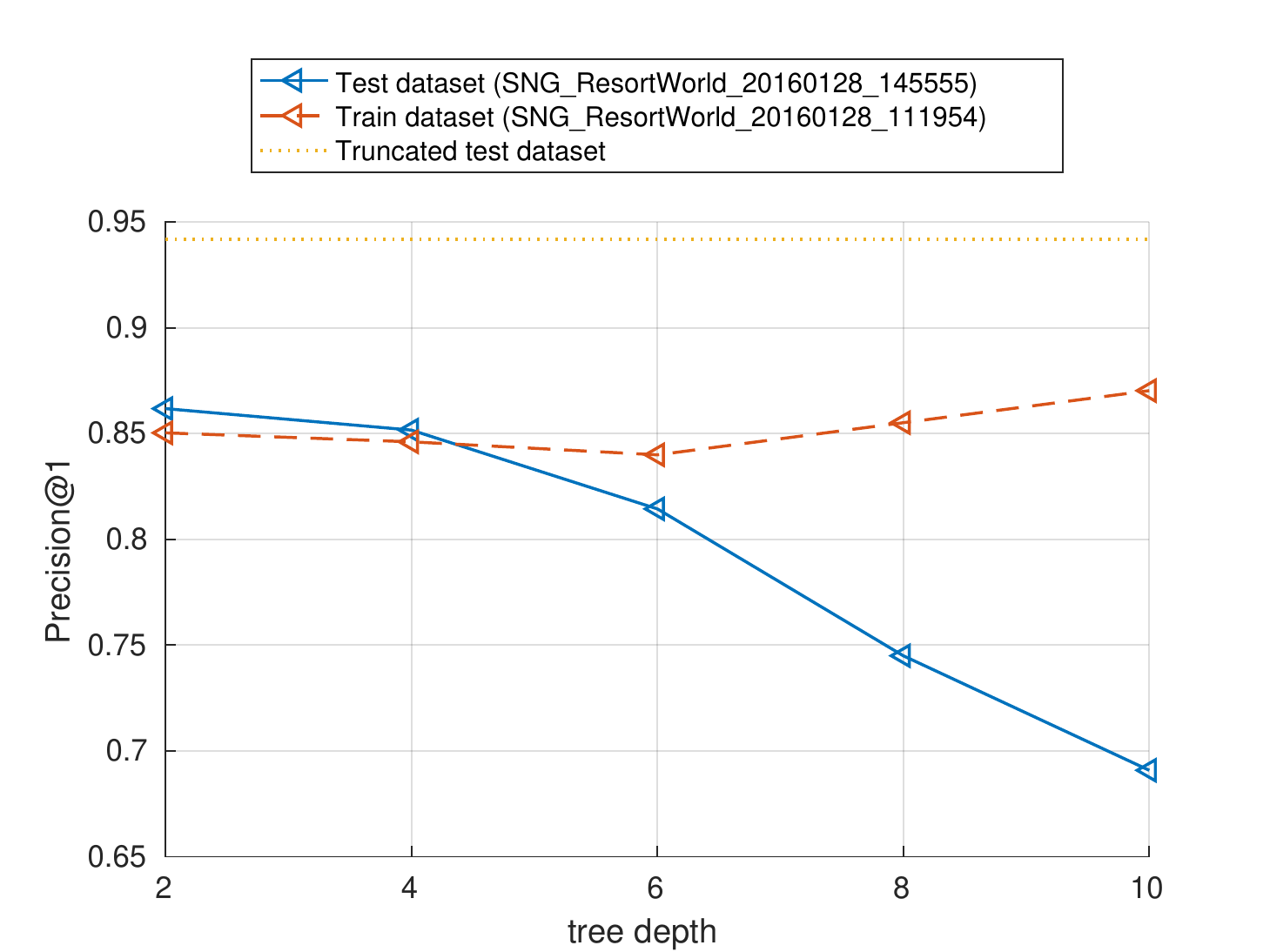}
	\label{jk:fig:fh1:a}	
	}
	\subfloat[256-bit codes][]{\includegraphics[width= 2.55in]{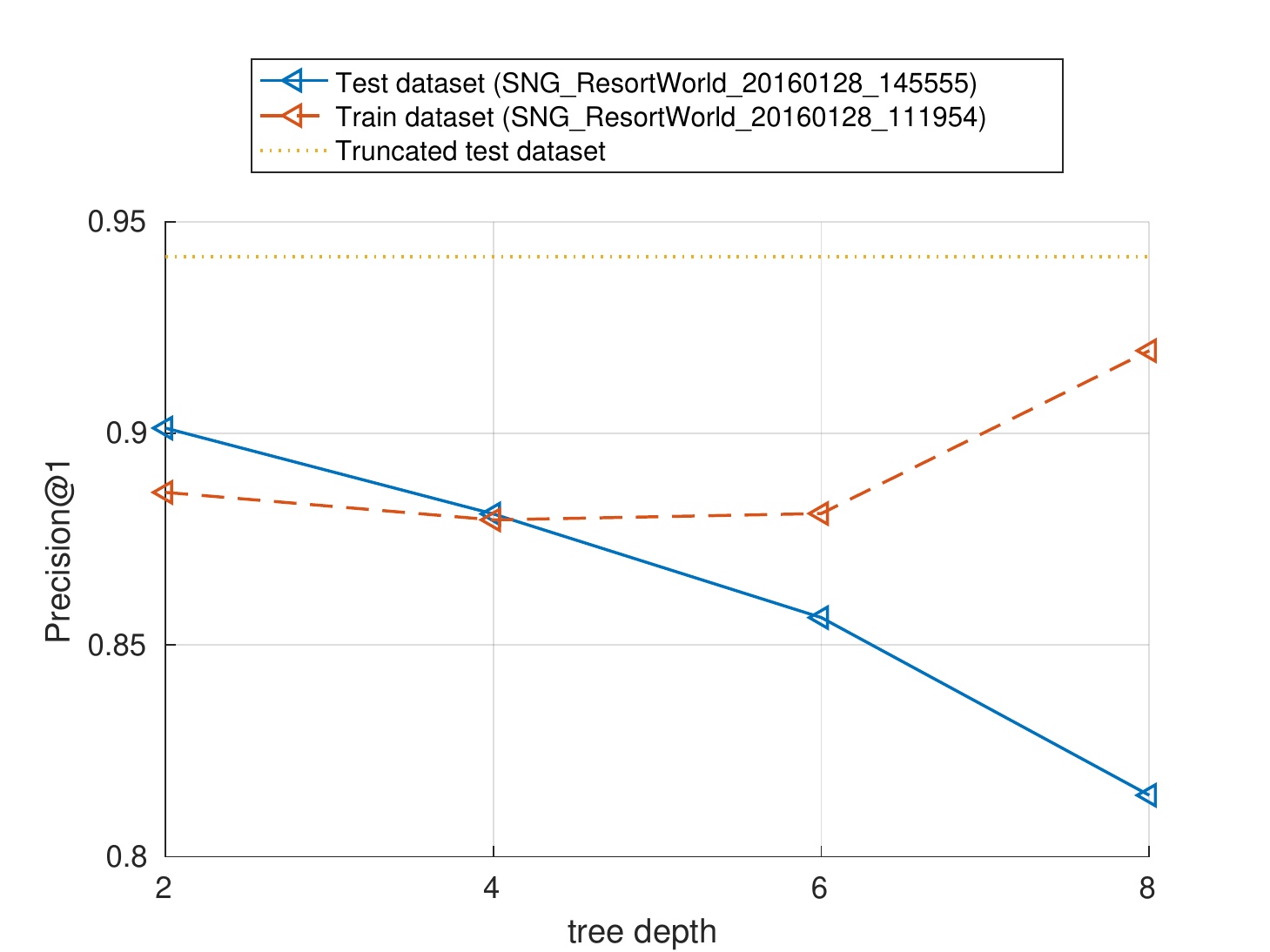}
	\label{jk:fig:fh1:b}	
	}
	\caption{Search precision using 256-bit hash codes generated from the test dataset by supervised FastHash method.
For each element from the training set, similarity information was encoded for 20 nearest neighbours, all similar and different number of dissimilar elements: 100 in \protect \subref{jk:fig:fh1:a} and 300 in \protect\subref{jk:fig:fh1:b}.
For comparison, search precision on the training dataset is shown with a dashed line and search precision in the test dataset truncated to first $k$ bits is plot with a dotted line.}
  	\label{jk:fig:fh1}
\end{figure*}

The key decision when using supervised hashing methods is a choice of similarity encoding scheme.
For large datasets it's not feasible to encode similarity between all $N \times N$ pairs of elements.
Figure \ref{jk:fig:fh1:a} presents results when for each element from the training dataset, similarity for its 20 nearest neighbours, all similar and 100 randomly chosen dissimilar elements is encoded.
Figure \ref{jk:fig:fh1:b} shows results when number of selected dissimilar elements is increased to 300.
In both cases search precision is worse compared to naive bit truncation.
It can be observed that encoding similarity information between more pairs improves the performance.
Unfortunately, for practical reasons, it was not possible to further increase amount of supervised training data.
For a training dataset consisting of over 300 thousand elements, encoding similarities between each element and over 300 other elements produces over 100 million pairs.
Learning procedure requires over 10 GB of memory to efficiently process such amount of data.

In both cases severe overfitting can be observed. As decision tree depth grows and model complexity increases, the discrepancy between performance on the test and training set grows.

\section{Conclusions and future work}

Evaluation of hashing methods on large datasets of binary FREAK descriptors yield some surprising results.
For 256-bit hash codes (half of the length of the original FREAK descriptor) none of the examined methods performed better compared to naive bit truncation approach.
In theory, linear projection-based hash methods should be able to produce at least as good result. Yet all of the evaluated methods performed worse.
For short codes, hashing methods gain advantage. 
For 128-bit codes Isotropic Hashing \cite{NIPS2012_4846}, the best unsupervised method, allows achieving 2\% better search accuracy compared to naive bit truncation. For 64-bit codes, Isotropic Hashing produces 15\% better results.

Examined semi-supervised and supervised methods were not able to benefit from the additional supervised information in the form of landmark ids linked with feature descriptors in the training set.
Surprisingly, adding more supervised information in the examined semi-supervised methods produced worse results. The likely reason is, that due to limited hardware resources, similarity information can be explicitly encoded only for fraction of pairs of points from the training dataset, which leads to suboptimal performance.

Hashing methods can be beneficial when storage is a primary concern and short codes are required.
They can be used to generate compact, 64-bit binary codes, form original 512-bit FREAK feature descriptors.
This would decrease storage requirements four times, and increase descriptor matching speed at the expense of  moderate decrease in search precision (about 15\%).
Further reduction to 32 bits has a detrimental effect on search precision, reducing it by over 40\%.

As a future work it will be beneficial to investigate other approaches to supervised dimensionality reduction, such as Mahalanobis metric learning.

\section*{Acknowledgment}
This research was supported by Google's Sponsor Research Agreement under the project "Efficient visual localization on mobile devices".

%

\bibliography{jk_bib}{}

\begin{thebibliography}{10}

\bibitem{6247715}
A.~Alahi, R.~Ortiz, and P.~Vandergheynst.
\newblock Freak: Fast retina keypoint.
\newblock In {\em 2012 IEEE Conference on Computer Vision and Pattern
  Recognition}, June 2012.

\bibitem{2016arXiv161207545C}
D.~{Cai}.
\newblock {A Revisit of Hashing Algorithms for Approximate Nearest Neighbor
  Search}.
\newblock {\em ArXiv e-prints}, December 2016.

\bibitem{6296665}
Y.~Gong, S.~Lazebnik, A.~Gordo, and F.~Perronnin.
\newblock Iterative quantization: A procrustean approach to learning binary
  codes for large-scale image retrieval.
\newblock {\em IEEE Transactions on Pattern Analysis and Machine Intelligence},
  35(12), Dec 2013.

\bibitem{6248024}
J.~P. Heo, Y.~Lee, J.~He, S.~F. Chang, and S.~E. Yoon.
\newblock Spherical hashing.
\newblock In {\em 2012 IEEE Conference on Computer Vision and Pattern
  Recognition}, June 2012.

\bibitem{Indyk:1998:ANN:276698.276876}
Piotr Indyk and Rajeev Motwani.
\newblock Approximate nearest neighbors: Towards removing the curse of
  dimensionality.
\newblock In {\em Proceedings of the Thirtieth Annual ACM Symposium on Theory
  of Computing}, STOC '98, New York, NY, USA, 1998. ACM.

\bibitem{NIPS2012_4846}
Weihao Kong and Wu~jun Li.
\newblock Isotropic hashing.
\newblock In F.~Pereira, C.~J.~C. Burges, L.~Bottou, and K.~Q. Weinberger,
  editors, {\em Advances in Neural Information Processing Systems 25}. Curran
  Associates, Inc., 2012.

\bibitem{NIPS2009_3667}
Brian Kulis and Trevor Darrell.
\newblock Learning to hash with binary reconstructive embeddings.
\newblock In Y.~Bengio, D.~Schuurmans, J.~D. Lafferty, C.~K.~I. Williams, and
  A.~Culotta, editors, {\em Advances in Neural Information Processing Systems
  22}. Curran Associates, Inc., 2009.

\bibitem{kulis2009kernelized}
Brian Kulis and Kristen Grauman.
\newblock Kernelized locality-sensitive hashing for scalable image search.
\newblock In {\em Computer Vision, 2009 IEEE 12th International Conference on}.
  IEEE, 2009.

\bibitem{FastHash2015Lin}
Guosheng Lin, Chunhua Shen, and Anton {van den Hengel}.
\newblock Supervised hashing using graph cuts and boosted decision trees.
\newblock {\em IEEE Transactions on Pattern Analysis and Machine Intelligence},
  2015.

\bibitem{lin2015supervised}
Guosheng Lin, Chunhua Shen, and Anton van~den Hengel.
\newblock Supervised hashing using graph cuts and boosted decision trees.
\newblock {\em IEEE transactions on pattern analysis and machine intelligence},
  37(11), 2015.

\bibitem{DBLP:journals/corr/abs-1205-2930}
Yue Lin, Deng Cai, and Cheng Li.
\newblock Density sensitive hashing.
\newblock {\em CoRR}, abs/1205.2930, 2012.

\bibitem{lin2013compressed}
Yue Lin, Rong Jin, Deng Cai, Shuicheng Yan, and Xuelong Li.
\newblock Compressed hashing.
\newblock In {\em Proceedings of the IEEE Conference on Computer Vision and
  Pattern Recognition}, 2013.

\bibitem{muja2009fast}
Marius Muja and David~G Lowe.
\newblock Fast approximate nearest neighbors with automatic algorithm
  configuration.
\newblock {\em VISAPP (1)}, 2(331-340), 2009.

\bibitem{Trzcinski20122173}
Tomasz Trzcinski, Vincent Lepetit, and Pascal Fua.
\newblock Thick boundaries in binary space and their influence on
  nearest-neighbor search.
\newblock {\em Pattern Recognition Letters}, 33(16), 2012.

\bibitem{WangSSJ14}
Jingdong Wang, Heng~Tao Shen, Jingkuan Song, and Jianqiu Ji.
\newblock Hashing for similarity search: {A} survey.
\newblock {\em CoRR}, abs/1408.2927, 2014.

\bibitem{DBLP:journals/corr/WangZSSS16}
Jingdong Wang, Ting Zhang, Jingkuan Song, Nicu Sebe, and Heng~Tao Shen.
\newblock A survey on learning to hash.
\newblock {\em CoRR}, abs/1606.00185, 2016.

\bibitem{WangSSH2012}
Jun Wang, Shih-Fu Chang, and S.~Kumar.
\newblock Semi-supervised hashing for large-scale search.
\newblock {\em IEEE Transactions on Pattern Analysis and Machine Intelligence},
  34(undefined), 2012.

\bibitem{icml2010_WangKC10}
Jun Wang, Sanjiv Kumar, and Shih fu~Chang.
\newblock Sequential projection learning for hashing with compact codes.
\newblock In Johannes Fürnkranz and Thorsten Joachims, editors, {\em
  Proceedings of the 27th International Conference on Machine Learning
  (ICML-10)}. Omnipress, 2010.

\bibitem{WangLKC15}
Jun Wang, Wei Liu, Sanjiv Kumar, and Shih{-}Fu Chang.
\newblock Learning to hash for indexing big data - {A} survey.
\newblock {\em CoRR}, abs/1509.05472, 2015.

\bibitem{NIPS2008_3383}
Yair Weiss, Antonio Torralba, and Rob Fergus.
\newblock Spectral hashing.
\newblock In D.~Koller, D.~Schuurmans, Y.~Bengio, and L.~Bottou, editors, {\em
  Advances in Neural Information Processing Systems 21}. Curran Associates,
  Inc., 2009.

\bibitem{journals/tkde/WuZCCB13}
Chenxia Wu, Jianke Zhu, Deng Cai, Chun Chen, and Jiajun Bu.
\newblock Semi-supervised nonlinear hashing using bootstrap sequential
  projection learning.
\newblock {\em IEEE Trans. Knowl. Data Eng.}, 25(6), 2013.

\bibitem{Xu:2013:HH:2540128.2540389}
Bin Xu, Jiajun Bu, Yue Lin, Chun Chen, Xiaofei He, and Deng Cai.
\newblock Harmonious hashing.
\newblock In {\em Proceedings of the Twenty-Third International Joint
  Conference on Artificial Intelligence}, IJCAI '13. AAAI Press, 2013.

\end{thebibliography}
\bibliographystyle{plain}



\end{document}